\documentclass[10pt,twocolumn,letterpaper]{article}

\usepackage[pagenumbers]{cvpr} 

\usepackage{bm}
\usepackage{tabularx}
\usepackage{xstring}
\usepackage{multirow}
\usepackage{xspace}
\usepackage{url}
\usepackage{subcaption}
\usepackage{xcolor}
\definecolor{cvprblue}{rgb}{0.21,0.49,0.74}
\usepackage[pagebackref,breaklinks,colorlinks,allcolors=cvprblue]{hyperref}

\def\paperID{*****} 
\def\confName{CVPR}
\def\confYear{2025}

\title{DELT: A Simple Diversity-driven \texttt{EarlyLate} Training for Dataset Distillation}

\author{     Zhiqiang Shen\thanks{Equal contribution. Work done while Ammar and Shitong visiting MBZUAI. Code is at: \url{https://github.com/VILA-Lab/DELT}.} \qquad
     Ammar Sherif\footnotemark[1] \qquad
     Zeyuan Yin  \qquad 
    Shitong Shao \\
    VILA Lab, MBZUAI  \\
    {\tt\small \{zhiqiang.shen, zeyuan.yin\}@mbzuai.ac.ae \tt\small \{ammarsherif90, 1090784053sst\}@gmail.com}
}

\begin{document}
\maketitle
\begin{abstract}
  Recent advances in dataset distillation have led to solutions in two main directions. The conventional { batch-to-batch} matching mechanism is ideal for small-scale datasets and includes bi-level optimization methods on models and syntheses, such as FRePo, RCIG, and RaT-BPTT, as well as other methods like distribution matching, gradient matching, and weight trajectory matching. Conversely, { batch-to-global} matching typifies decoupled methods, which are particularly advantageous for large-scale datasets. This approach has garnered substantial interest within the community, as seen in SRe$^2$L, G-VBSM, WMDD, and CDA. A primary challenge with the second approach is the lack of diversity among syntheses within each class since samples are optimized independently and the same global supervision signals are reused across different synthetic images. In this study, we propose a new {\bf\em D}iversity-driven \texttt{{\bf\em E}arly{\bf\em L}ate} {\bf\em T}raining (DELT) scheme to enhance the diversity of images in {batch-to-global} matching with less computation. Our approach is conceptually simple yet effective, it partitions predefined IPC samples into smaller subtasks and employs local optimizations to distill each subset into distributions from distinct phases, reducing the uniformity induced by the unified optimization process. These distilled images from the subtasks demonstrate effective generalization when applied to the entire task. We conduct extensive experiments on CIFAR, Tiny-ImageNet, ImageNet-1K, and its sub-datasets. Our approach outperforms the previous state-of-the-art by 2$\sim$5\% on average across different datasets and IPCs (images per class), increasing diversity per class by more than 5\% while reducing synthesis time by up to 39.3\% for enhancing the training efficiency.
\end{abstract}    
\section{Introduction}
\label{sec:intro}

\begin{figure}
  \centering
    \includegraphics[width=0.95\linewidth]{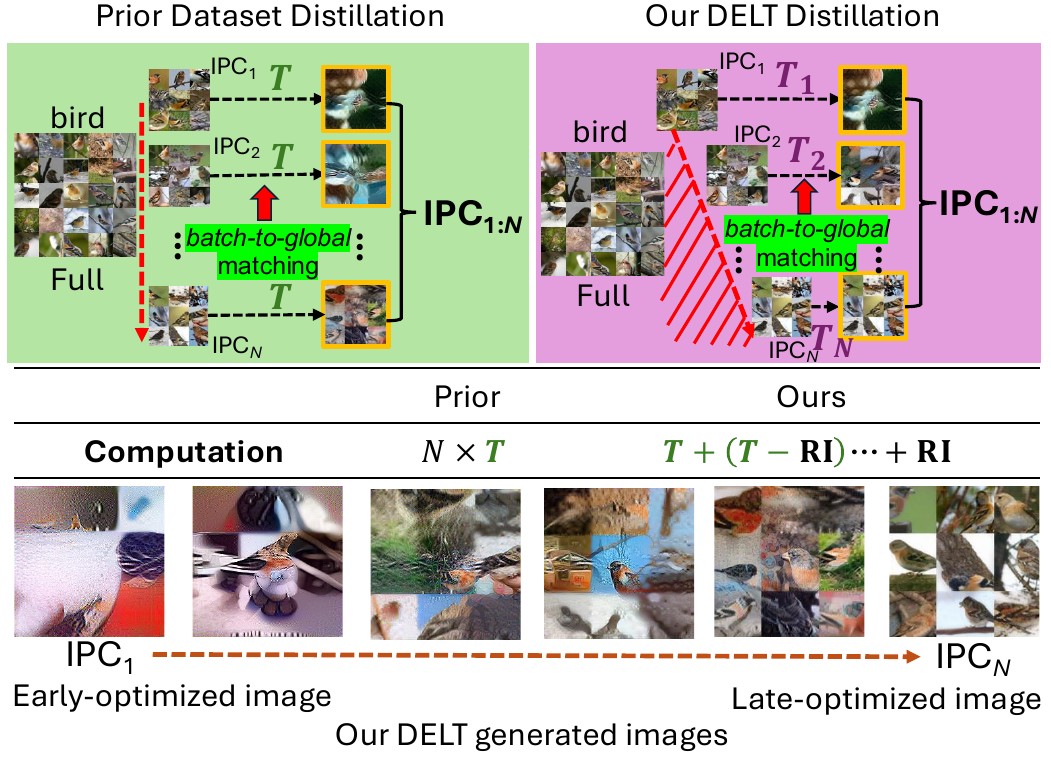}
  \vspace{-0.15in}
  \caption{Distilling datasets to IPC$_{N}$ requires $N\!\times \!T$ iterations in traditional distillation processes (left) but fewer iterations in our \texttt{EarlyLate} strategy (right). IPC$_{1:N}$ represents a set of images from 1 to $N$. The red shaded area is our saved computational cost.}
  \label{fig:motivation}
   \vspace{-0.15in}
\end{figure}

\begin{figure*}[t]
  \centering
  \includegraphics[width=0.99\linewidth]{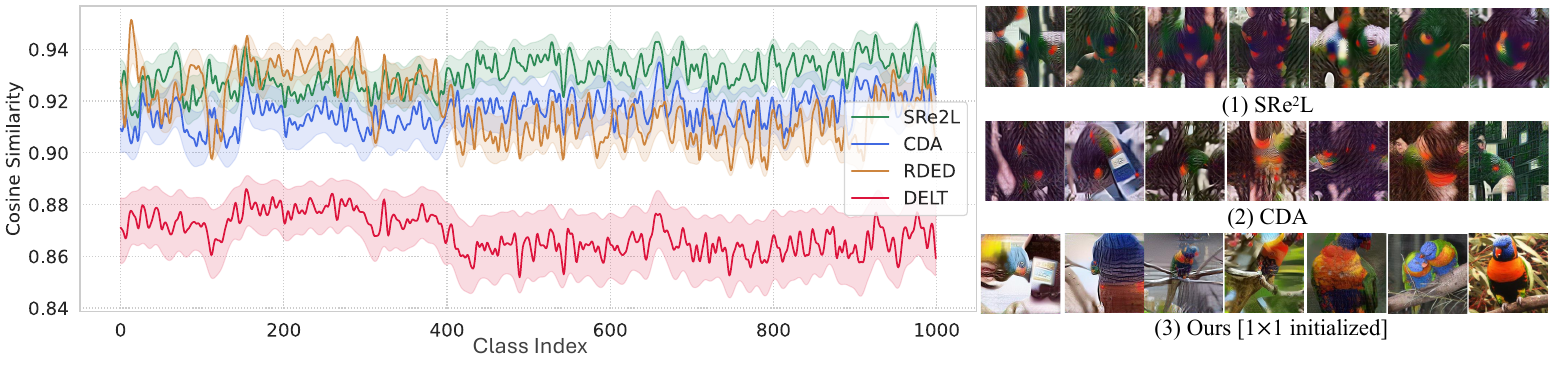}
  \vspace{-0.1in}
  \caption{{\bf Left}: Intra-class semantic cosine similarity after a pretrained ResNet-18 model on ImageNet-1K dataset, lower values are better. {\bf Right}: Synthetic images from SRe$^2$L, CDA and our DELT.}
  \label{fig:diversity}
   \vspace{-0.1in}
\end{figure*}

In the era of large models and large datasets, dataset distillation has emerged as a crucial strategy to enhance training efficiency and make advanced technologies more accessible and affordable for the general public. Previous approaches~\cite{wang2018dataset,zhao2020dataset,zhou2022dataset,cazenavette2022dataset,lee2022dataset,DBLP:conf/wacv/ZhaoB23,liu2022dataset,cui2023scaling,chen2024data,he2024multisize} primarily employ a {\em batch-to-batch} matching technique, where information like features, gradients, and trajectories from a local original data batch are used to supervise and train a corresponding batch of generated data. The strength of this method lies in its ability to capture fine-grained information from the original data, as each batch's supervision signals vary. However, the downside is the necessity to repeatedly input both original and generated data for each training iteration, which significantly increases memory usage and computational costs. Recently, a new decoupled method~\cite{yin2023squeeze,yin2023dataset,liu2023dataset} has been proposed to separate the model training and data synthesis, also it leverages the {\em batch-to-global} matching to avoid inputting original data during distilled data generation. This solution has demonstrated great advantage on large-scale datasets like ImageNet-1K~\cite{yin2023squeeze,shao2023generalized} and ImageNet-21K~\cite{yin2023dataset}. However, as shown in Fig.~\ref{fig:diversity}, a significant limitation of this approach is the lack of diversity caused by the mechanism of synthesizing each data point individually, where supervision is repetitively applied across various synthetic images. For instance, SRe$^2$L~\cite{yin2023squeeze} utilizes globally-counted layer-wise running means and variances from the pre-trained model for supervising different intra-class image synthesis. This methodology results in a severely limited diversity within the same category of generated images.

To address this issue, a few prior studies~\cite{shao2023generalized,sun2024diversity} have proposed to enlarge diversity within each class. For instance, G-VBSM~\cite{shao2023generalized} utilizes a diverse set of {\em local-match-global} matching signals derived from multiple backbones and statistical metrics, to achieve more precise and effective matching than the singular model. However, as the diversity of matching models grows, the overall complexity of the framework also increases thus diminishing its conciseness.
RDED~\cite{sun2024diversity} crops each original image into multiple patches and ranks these using realism scores generated by an observer model. Then it stitches every four chosen patches from previous stage into a single new image to produce IPC-numbered distilled images for each class. RDED is efficient to combine multiple images but does not enhance or optimize the visual content on the distilled dataset, thus the diversity and richness of information are largely dependent on the distribution of the original dataset. 

Our solution, termed the \texttt{EarlyLate} training scheme, is straightforward and also orthogonal to these prior methods: by initializing each image in the same category at a different starting point for optimization, we ensure that the final optimized results vary significantly across images. We also use teacher-ranked real image patches to initialize the synthetic images. This prevents some images from being short-optimized and ensures they provide sufficient information. As shown in Fig.~\ref{fig:motivation} of the computation comparison, our approach not only enhances intra-class diversity but also dramatically reduces the computational load of the training process by 39.3\% on ImageNet-1K. Specifically, while conventional training requires $T$ optimization iterations per image or batch, in our \texttt{EarlyLate} scheme, the first image undergoes $T_1$ iterations (where $T_1=T$). Subsequent batches are processed with progressively fewer iterations, such as $T_2$ ($T_2=T_1-\text{RI}$\footnote{{RI} is the number of round iterations and will be introduced in Sec.~\ref{ablation}.}) for the next set, and so forth. The iterations for the final batch are reduced to {RI} which is $1/j$ of the standard count (where typically $j=$ 4 or 8), meaning the total number of optimization iterations required is just about $2/3$ of prior {\em batch-to-global} matching methods, such as SRe$^2$L and CDA. We further visualize the average cosine similarity between each sample of 50 IPCs using the associated cluster centroid within the same class on ImageNet-1K, as shown in Fig.~\ref{fig:diversity} left subfigure, our DELT illustrates a smaller similarity significantly, and also shows substantially better visual diversity than other counterparts across all classes, as in the right subfigure of Fig.~\ref{fig:diversity}.

We conduct extensive experiments on various datasets of CIFAR-10, Tiny-ImageNet, ImageNet-1K and its subsets. On ImageNet-1K, our proposed approach achieves 66.1\% under IPC 50 with ResNet-101, outperforming previous state-of-the-art RDED by 4.9\%. On small-scale datasets of CIFAR-10, our approach also obtains 2.5\% and 19.2\% improvement over RDED and SRe$^2$L using ResNet-101. 

Our main contributions in this work are as follows:

\begin{itemize}
\addtolength{\itemsep}{-0.0in}
\item We propose a simple yet effective \texttt{EarlyLate} training scheme for dataset distillation to enhance intra-class diversity of synthetic images for {\em batch-to-global} matching.

\item We demonstrate empirically that the proposed method can generate optimized images at different distances with a fast speed, to enlarge informativeness among generations.

\item We conducted extensive experiments and ablations on various datasets across different scales to prove the effectiveness of the proposed approach\footnote{{Our synthetic images on ImageNet-1K are available at \href{https://drive.google.com/file/d/1Rr_ik94FNte75yc4GiKtv927qdBwKskr/view?usp=sharing}{link}}.}.

\end{itemize}

\section{Related Work}
\label{sec:related}

\noindent{\bf Dataset Distillation.} Dataset distillation or condensation~\cite{wang2018dataset} focuses on creating a compact yet representative subset from a large original dataset. This enables more efficient model training while maintaining the ability to evaluate on the original test data distribution and achieve satisfactory performance. Previous works~\cite{wang2018dataset,zhao2020dataset,zhou2022dataset,cazenavette2022dataset,lee2022dataset,DBLP:conf/wacv/ZhaoB23,liu2022dataset,cui2023scaling,chen2024data,he2024multisize} mainly designed how to better match the distribution between original data and generated data in a {\em batch-to-batch} manner, such as the distribution of features~\cite{DBLP:conf/wacv/ZhaoB23}, gradients~\cite{zhao2020dataset}, or the model weight trajectories~\cite{cazenavette2022dataset,cui2023scaling}. The primary optimization method used is bi-level optimization~\cite{liu2021investigating,zhang2023introduction}, which involves optimizing model parameters and updating images simultaneously. For instance, using gradient matching, the process can be formulated as to minimize the gradient distance:
\vspace{-0.05in}
\begin{equation}
\begin{aligned}
&\min _{\mathcal{S} \in \mathbb{R}^{N \times d}} D\left(\nabla_\theta \ell(\mathcal{S} ; \theta), \nabla_\theta \ell(\mathcal{T} ; \theta)\right)=D(\mathcal{S}, \mathcal{T} ; \theta),
\end{aligned}
\vspace{-0.05in}
\end{equation}
where the function $D(\cdot,\cdot)$ is defined as a distance metric such as MSE~\cite{wang2009mean}, $\theta$ denotes the model parameters, and $\nabla_\theta \ell(\cdot;\theta)$ represents the gradient, utilizing either the original dataset $\mathcal{T}$ or its synthetic version $\mathcal{S}$. $N$ is the number of $d$-dimensional synthetic data. During distillation, the synthetic dataset $\mathcal{S}$ and model $\theta$ are updated alternatively,
\begin{equation}
\mathcal{S} \leftarrow \mathcal{S}-\lambda \nabla_{\mathcal{S}} D(\mathcal{S}, \mathcal{T} ; \theta), \quad \theta \leftarrow \theta-\eta \nabla_\theta \ell(\theta ; \mathcal{S}).
\end{equation}
where $\lambda$ and $\eta$ are learning rates designated for $\mathcal{S}$ and $\theta$.

\begin{figure}[t]
  \centering
    \includegraphics[width=0.86\linewidth]{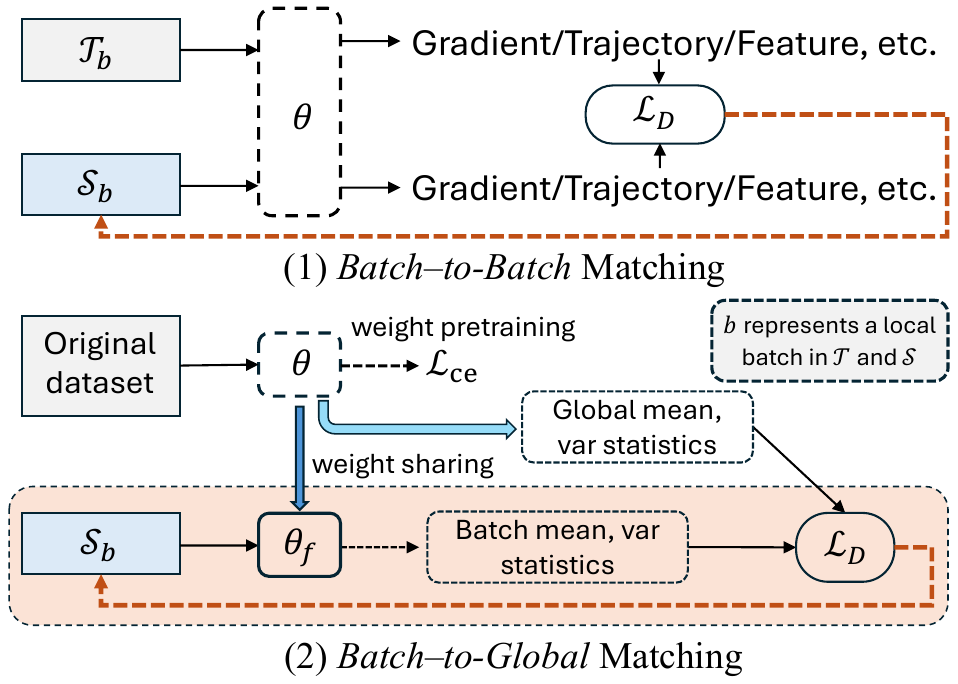}
  \vspace{-0.1in}
  \caption{{\em Batch–to-batch} vs. {\em batch-to-global} matching in DD. $\theta_f$ indicates weights are pretrained and frozen in synthesis stage.}
  \label{fig:comparison}
   \vspace{-0.15in}
\end{figure}

\noindent{\bf Diversity in Dataset Distillation.}
{\em Batch-to-global} matching used in~\cite{yin2023squeeze,shao2023generalized,yin2023dataset,liu2023dataset,xiao2024large,du2024diversity} tracks the distribution of BN statistics derived from original dataset for the local batch synthetic data. However, this type of approach can easily encounter diversity issues within the same class due to the optimization objective.
Fig.~\ref{fig:comparison} illustrates the difference of {\em batch-to-batch} and {\em batch-to-global} matching mechanisms, where $b$ represents a local batch in data $\mathcal{T}$ and $\mathcal{S}$.

Moreover, for the recent advances of multi-stage dataset distillation methods, MDC~\cite{he2024multisize} proposes to compress multiple condensation processes into a single one by including an adaptive subset loss on top of the basic condensation loss, so that to obtain datasets with multiple sizes. PDD~\cite{chen2024data} generates multiple small batches of synthetic images, each batch is conditioned on the accumulated data from previous batches. Unlike PDD, our current synthetic batch is independent with different operation iterations and not relevant to any previous batches. D3~\cite{qin2024distributional} partitions large datasets into smaller subtasks and employs locally trained experts to distill each subset into distributions. These distilled distributions from the subtasks demonstrate effective generalization when applied to the entire task. The recently proposed LPLD~\cite{xiao2024large} batches images by class, leveraging the natural independence between classes, and introduces class-wise supervision for alignment. 

\section{Approach}
\label{sec:approach}

\noindent{\bf Preliminaries.} 
The objective of a regular dataset distillation task is to generate a compact synthetic dataset $\mathcal{S} = \{\left(\bm{\hat x}_{1}, \bm{\hat y}_{1}\right), \ldots, \left(\bm{\hat x}_{|\mathcal{S}|}, \bm{\hat y}_{|\mathcal{S}|}\right)\}$ as 
a {\em student} dataset that captures a substantial amount of the information from a larger labeled dataset $\mathcal{T}=\{\left(\bm{x}_{1}, \bm{y}_{1}\right), \ldots, \left(\bm{x}_{|\mathcal{T}|}, \bm{y}_{|\mathcal{T}|}\right)\}$, which serves as the {\em teacher} dataset. Here, $\bm{\hat y}$ represents the soft label for the synthetic sample $\bm{\hat x}$, and the size of $\mathcal{S}$ is much smaller than $\mathcal{T}$, yet it retains the essential information of the original dataset $\mathcal{T}$. The learning goal using this distilled data is to train a post-validation model with parameters $\boldsymbol{\theta}$:
\begin{equation}
\boldsymbol{\theta}_{\mathcal{S}} = \underset{\boldsymbol{\theta}}{\arg \min } \mathcal{L}_{\mathcal{S}}(\boldsymbol{\theta}),
\end{equation}
\begin{equation}
\mathcal{L}_{\mathcal{S}}(\boldsymbol{\theta}) = \mathbb{E}_{(\bm{\hat x}, \bm{\hat y}) \in \mathcal{S}} \left[\ell(\phi_{\boldsymbol{\theta}_{\mathcal{S}}}(\bm{\hat x}), \bm{\hat y}; \boldsymbol{\theta})\right],
\end{equation}
where $\ell$ is a standard loss function such as soft cross-entropy and $\phi_{\boldsymbol{\theta}_{\mathcal{S}}}$ represents the model.

The primary aim of dataset distillation is to produce synthetic data that ensures minimal performance difference between models trained on the synthetic dataset $\mathcal{S}$ and those trained on the original dataset $\mathcal{T}$ using validation data ${V}$. The optimization procedure for generating $\mathcal{S}$ is given by:
\begin{equation}
\begin{aligned}
\underset{\mathcal{S},|\mathcal{S}|}{\arg \min }&\left(\operatorname { s u p } \left\{\mid \ell\left(\phi_{\theta_{\mathcal{T}}}\left(x_{\text {val }}\right), y_{\text {val }}\right)\right.\right. \\
& -\ell\left(\phi_{\theta_{\mathcal{S}}}\left(x_{v a l}\right), y_{v a l}\right)_{\left(x_{v a l}, y_{v a l}\right) \sim V} .
\end{aligned}
\end{equation}
where $({\bm x}_{val}, {\bm y}_{val})$ are the sample and label pairs in the validation set of the real dataset $\mathcal{T}$. The learning task then focuses on the $<$data, label$>$ pairs within $\mathcal{S}$, maintaining a balanced representation of distilled data across each class.

\begin{figure*}[t]
  \centering
  \includegraphics[width=0.9\linewidth]{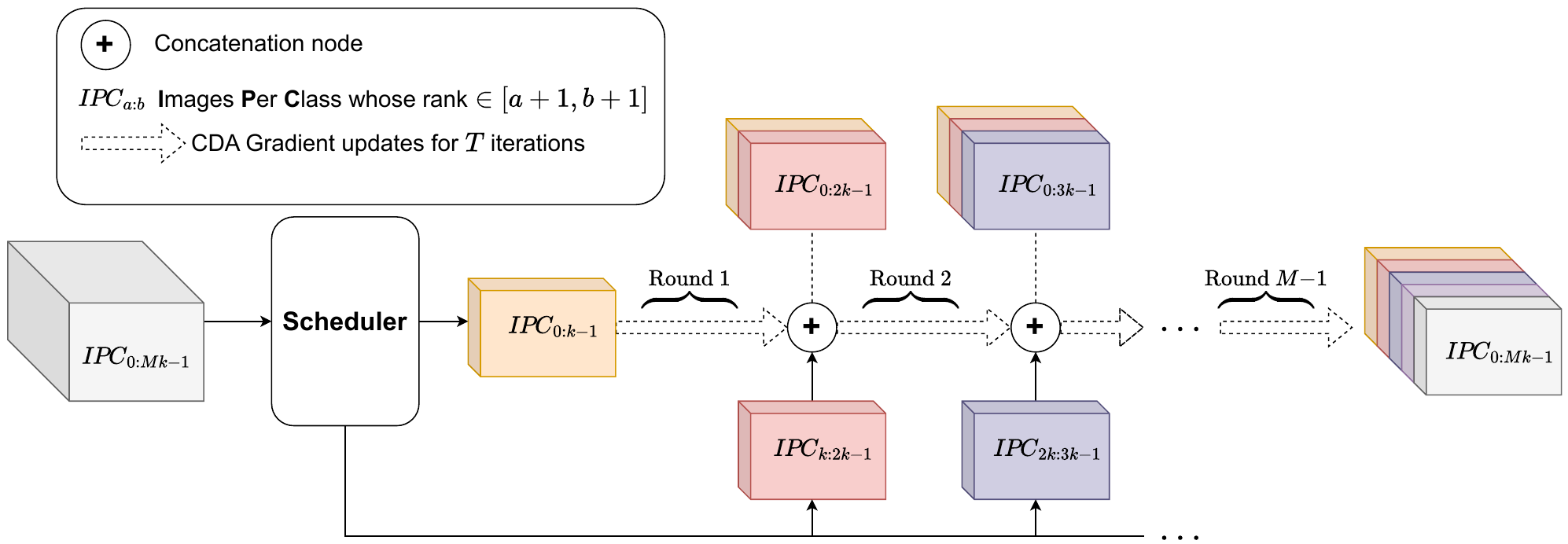}
  \vspace{-0.1in}
  \caption{The proposed DELT learning procedure via a multi-round {\texttt{EarlyLate}} scheme.}
  \label{fig:delt_learning_procedure}
   \vspace{-0.18in}
\end{figure*}

\noindent{\bf Initialization.} 
Previous dataset distillation methods~\cite{yin2023squeeze,shao2023generalized,yin2023dataset} on large-scale datasets like ImageNet-1K and 21K employ Gaussian noise by default for data initialization in the synthesis phase. However, Gaussian noise is random and lacks any semantic information. Intuitively, using real images provides a more meaningful and structured starting point, and this structured start can lead to quicker convergence during optimization because the initial data already contains useful features and patterns that are closer to the target distribution, which further enhances realism, quality, and generalization of the synthesized images. As shown in Fig.~\ref{fig:diversity} right subfigure, our generated images exhibit both diversity and a high degree of realism in some cases.

\noindent{\bf Selection Criteria.} Here, we introduce how to select real image patches to initialize the synthetic images. In our final syntheses, a significant fraction of our data has been subject to limited optimization iterations, making effective initialization crucial. A proper initialization also dramatically minimizes the overall computational load required for the updating degree on data. Prior approach~\cite{sun2024diversity} has demonstrated that choosing representative data patches from the original dataset without training can yield favorable performance without any additional training. Our observation, however, underscores that applying iterative refinement to original patches can lead to markedly improved results. 

\begin{figure}[h]
  \vspace{-0.1in}
  \centering
    \includegraphics[width=0.82\linewidth]{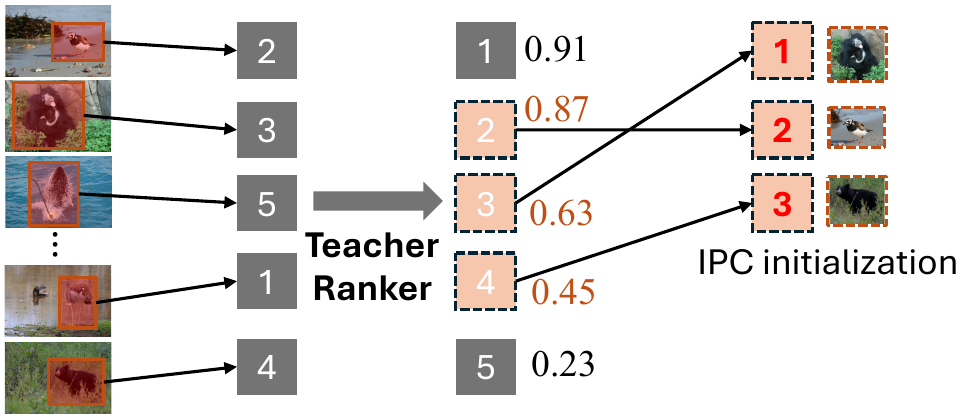}
  \vspace{-0.15in}
  \caption{Selection criteria with a teach ranker.}
  \label{fig:selection}
   \vspace{-0.15in}
\end{figure}

As illustrated in Fig.~\ref{fig:selection}, our selection criterion is based on a pretrained teacher model as a ranker, we calculate all patches' probabilities and sort them as the initialization pool. Then, we choose a patch of images scoring closer to the per-class median as initialization for optimization. More details regarding initialization and order can be found in Appendix.
The motivation is that such images have a medium difficulty level to the teacher, so they have more room for information enhancement via distillation gradients. We further empirically validate this strategy by comparing different strategies in Table~\ref{ablation-selection-criteria}.

\noindent{\bf Diversity-driven IPC Concatenation Training.} 
As shown in Fig.~\ref{fig:delt_learning_procedure}, to further emphasize diversity and avoid potential distribution bias from initialization, we optimize the initialized images starting from different points. The motivation behind this design is that different data samples require varying numbers of iterations to converge as the early stopping~\cite{prechelt2002early} from other research domain. Importantly, as images become easier to predict with more updates by class labels, training primarily on easy data points can hinder model generalization. Therefore, our method enhances generalization by generating data samples with varying difficulty levels, acting as a regularizer by limiting the optimization process to a smaller volume of image pixel space. Previous work~\cite{bai2021understanding} studies how to perform simple early stopping on different layers' weights with progressive retraining to mitigate noisy labels. Unlike it, we are pioneering to study both {\em early} and {\em late} training when optimizing data. Moreover, we improve the efficiency of our approach by performing gradient updates in a single scan. Initially, we conduct a single gradient loop, continually introducing new data for distillation by concatenating them at different time stamps.
Consequently, the $M$ batch receives the synthetic images of all preceding batches, $\text{\em IPC}_{0:M{k-1}}$, as final generations. This process can be simplified as follows: 
\vspace{-0.05in}
\begin{equation}
\text{\em IPC}_{0:M{k-1}}=[\underbrace{\underbrace{\underbrace{\bm{\hat x}_0, \bm{\hat x}_1, \ldots, \bm{\hat x}_{k-1}}_{\text{\em IPC}_{0:k-1}}, \ldots}_{\ldots}, \bm{\hat x}_{Mk-1}}_{\text{\em IPC}_{0:Mk-1}}]
\vspace{-0.05in}
\end{equation} 
where $\left[\bm{\hat x}_0, \bm{\hat x}_1, \ldots, \bm{\hat x}_{Mk-1}\right]$ refers to the concatenation of generated images. $M$ is the number of batches, $k$ is the number of generated images in each batch. 
We train these different batches at different starting points, each batch goes through a complete learning phase, but the total number of iterations varies. Then, the multiple IPCs of $\bm{\hat x}$ are concatenated into a simple batch. Because of its early-late training property, we refer to this scheme as \texttt{EarlyLate} training. 

\noindent{\bf Data synthesis.}  Our \texttt{EarlyLate} optimization procedure can be formulated as a multi-stage training scheme:  
\vspace{-0.05in}
\begin{equation} \label{objective}
\begin{aligned}
\mathrm{Round \ 1}\!: & \underset{\mathcal{C}_{\mathrm{IPC}_{0:k-1}},|\mathcal{C}|}{\arg \min } \ell\left(\phi_{\boldsymbol{\theta}_{\mathcal{T}}}\left(\widetilde{\boldsymbol{x}}_{\mathrm{IPC}_{0:k-1}}\right), \boldsymbol{y}\right)+\mathcal{R}_{\text {reg }} \\ 
... \\
\mathrm{Round \ M\!-\!1}\!: & \underset{\mathcal{C}_{\mathrm{IPC}_{0:Mk-1}},|\mathcal{C}|}{\arg \min } \ell\left(\phi_{\boldsymbol{\theta}_{\mathcal{T}}}\left(\widetilde{\boldsymbol{x}}_{\mathrm{IPC}_{0:Mk-1}}\right), \boldsymbol{y}\right)\!+\!\mathcal{R}_{\text {reg }}
\vspace{-0.05in}
\end{aligned}
\end{equation}
where $\mathcal{C}$ is the target distilled dataset. The number of batches $\mathrm{M}>1$ (If $\mathrm{M}=1$, training will degenerate into a way without \texttt{EarlyLate}). This process is referred to in Fig.~\ref{fig:delt_learning_procedure} bottom row. $\mathcal{R}_{\text {reg }}$ is the regularization term, we also utilize the BatchNorm distribution regularization term as in SRe$^2$L~\cite{yin2023squeeze} to improve the quality of the generated images.

\noindent{\bf Training Procedure.} Regarding concatenation training, we elaborate: Our \texttt{EarlyLate} enhances the diversity of the synthetic data by varying the number of iterations for different IPCs during data synthesis phase. This means the first IPC can be recovered for the largest number of iterations like 4K while the last IPC will only be recovered using 500 iterations.
To make this process efficient, we share the recovery time (on the GPU) across the different IPCs via concatenation to minimize time as much as possible.
Therefore, the first image IPC will start recovery for a couple of iterations, and when it completes iteration 3,500 the last IPC will join it in the recovery phase to get its 500 iterations.

 As illustrated in Fig.~\ref{fig:delt_learning_procedure}, our learning procedure is extremely simple using an incremental learning process: We split the total IPCs to be learned into multiple batches. The training begins with the first batch. Following a predefined number of iterations, the second batch commences its iterative training, and this process continues sequentially with subsequent batches. We define two types of optimization iterations for training: maximum iteration (MI) for the earliest batch training and round iteration (RI). MI presents the number of optimization iterations that the earliest batch goes through, i.e., the maximum number of iterations for the first batch's gradient updating. RI represents the number of iterations used for each round in Fig.~\ref{fig:delt_learning_procedure}. It essentially indicates the iteration gap between the optimization of two adjacent batches.
 {\em Batch-to-global} matching algorithm~\cite{yin2023dataset} of Eq.~\ref{objective}  is utilized between each round. In our DELT, later sub-batches will join the previous sub-batches in the image recovery stage instead of freezing the earlier sub-batches.

\vspace{-0.05in}
\section{Experiments}
\label{sec:exp}

\subsection{Datasets and Result Details}

We first run DELT on five standard benchmark tests including CIFAR-10 (10 classes)~\cite{krizhevsky2009learning}, Tiny-ImageNet (200 classes)~\cite{le2015tiny}, ImageNet-1K (1,000 classes)~\cite{deng2009imagenet} and it variants of ImageNette (10 classes)~\cite{Fastai}, and ImageNet-100 (100 classes)~\cite{tian2020contrastive} with performances reported in Table~\ref{tb:resnet} and Table~\ref{tb:main}. The evaluation protocol follows prior works~\cite{sun2024diversity,yin2023squeeze}. We compare DELT to six baseline dataset distillation algorithms including Matching Training Trajectories (MTT)~\cite{cazenavette2022dataset}, Improved Distribution Matching (IDM)~\cite{zhao2023improved}, TrajEctory Matching with Constant Memory (TESLA)~\cite{cui2023scaling}, Squeeze-Recover-Relabel (SRe$^2$L)~\cite{yin2023squeeze}, Difficulty-Aligned Trajectory-Matching (DATM)~\cite{guo2024lossless}, Realistic-Diverse-Efficient Dataset Distillation (RDED)~\cite{sun2024diversity}. Following previous dataset distillation methods~\cite{zhao2020dataset,sun2024diversity,yin2023squeeze}, we use ConvNet~\cite{gidaris2018dynamic}, ResNet-18/ResNet-101~\cite{he2016deep}, EfficientNet-B0~\cite{tan2019efficientnet}, MobileNet-V2~\cite{sandler2018mobilenetv2}, MnasNet1\_3~\cite{tan2019mnasnet}, and RegNet-Y-8GF~\cite{radosavovic2020designing}, as our backbone for training or post-validation. All our experiments are conducted on 4$\times$ NVIDIA RTX 4090 GPUs.

As shown in Table~\ref{tb:resnet}, our approach establishes the new state-of-the-art accuracy in {13} out of {15} of the configurations on five datasets from small-scale CIFAR-10 to large-scale ImageNet-1K using either relatively large backbone architecture of ResNet-101 or small MobileNet-v2, in many cases with significant margins of improvement. The results using small-scale architecture ConvNet are shown in Table~\ref{tb:main}, our approach also achieves the state-of-the-art accuracy in {7} out of {9} of the configurations on four datasets.

\begin{table*}[t]
    \centering
    \scalebox{0.82}{
        \begin{tabular}{@{}cc|ccccccccc}
            \toprule[1.4pt]
                         &     & \multicolumn{3}{c}{ResNet-18} & \multicolumn{3}{c}{ResNet-101}     &   \multicolumn{2}{c}{MobileNet-v2}            \\
                                           \cmidrule(lr){3-5} \cmidrule(lr){6-8} \cmidrule(lr){9-10}
            Dataset                        & IPC & SRe$^2$L~\cite{yin2023squeeze}   & RDED~\cite{sun2024diversity}  &   DELT (Ours)                  & SRe$^2$L~\cite{yin2023squeeze}  & RDED~\cite{sun2024diversity} &  DELT (Ours)                     &   RDED~\cite{sun2024diversity}  &  DELT (Ours)                    \\  \midrule
                                           & 1   & 16.6 $\pm$ 0.9                & 22.9 $\pm$ 0.4                & \textbf{24.0 $\pm$ 0.8}& 13.7 $\pm$ 0.2                  & 18.7 $\pm$ 0.1               &\ \textbf{20.4 $\pm$ 1.0} &  18.1 $\pm$ 0.9  & \textbf{20.2 $\pm$ 0.4}   \\
            CIFAR-10                       & 10  & 29.3 $\pm$ 0.5                   & 37.1 $\pm$ 0.3                & \textbf{43.0 $\pm$ 0.9}& 24.3 $\pm$ 0.6                  & 33.7 $\pm$ 0.3               &\ \textbf{37.4 $\pm$ 1.2} &    29.2 $\pm$ 1.1    &  \textbf{29.3 $\pm$ 0.3}   \\
                                           & 50  & 45.0 $\pm$ 0.7                   & 62.1 $\pm$ 0.1                & \textbf{64.9 $\pm$ 0.9}& 34.9 $\pm$ 0.1                  & 51.6 $\pm$ 0.4               &\ \textbf{54.1 $\pm$ 0.8} &     39.9 $\pm$ 0.5   & \textbf{42.9 $\pm$ 2.2}   \\ \midrule
                                           & 1   & 19.1 $\pm$ 1.1                   & \textbf{35.8 $\pm$ 1.0}       & 24.1 $\pm$ 1.8         & 15.8 $\pm$ 0.6                  & \textbf{25.1 $\pm$ 2.7}      & 19.4 $\pm$ 1.7           &   \textbf{26.4 $\pm$ 3.4}     &   {19.1 $\pm$ 1.0}   \\ 
            ImageNette                     & 10  & 29.4 $\pm$ 3.0                   & 61.4 $\pm$ 0.4                & \textbf{66.0 $\pm$ 1.4}& 23.4 $\pm$ 0.8                  & 54.0 $\pm$ 0.4               &\ \textbf{55.4 $\pm$ 6.2} &    52.7 $\pm$ 6.6    &   \textbf{64.7 $\pm$ 1.4}   \\
                                           & 50  & 40.9 $\pm$ 0.3                   & 80.4 $\pm$ 0.4                & \textbf{88.2 $\pm$ 1.2}& 36.5 $\pm$ 0.7                  & 75.0 $\pm$ 1.2               &\ \textbf{83.3 $\pm$ 1.1} &     80.0 $\pm$ 0.0    &   \textbf{85.7 $\pm$ 0.4}   \\ \midrule
            \multirow{3}{*}{Tiny-ImageNet} & 1   & 2.62 $\pm$ 0.1                   & \textbf{9.7 $\pm$ 0.4}        & 9.3 $\pm$ 0.5          & 1.9 $\pm$ 0.1                   & 3.8 $\pm$ 0.1                &\ \textbf{5.6 $\pm$ 1.0}  &      3.5 $\pm$ 0.1        &   \textbf{3.5 $\pm$ 0.5}   \\ 
                                           & 10  & 16.1 $\pm$ 0.2                   & 41.9 $\pm$ 0.2                &\textbf{43.0 $\pm$ 0.1} & 14.6 $\pm$ 1.1                  & 22.9 $\pm$ 3.3               &\ \textbf{42.8 $\pm$ 0.9} &     24.6 $\pm$ 0.1      &   \textbf{26.5 $\pm$ 0.5}   \\
                                           & 50  & 41.1 $\pm$ 0.4                   & \textbf{58.2 $\pm$ 0.1}       & 55.7 $\pm$ 0.5         & 42.5 $\pm$ 0.2                  & 41.2 $\pm$ 0.4               &\ \textbf{58.5 $\pm$ 0.3} &      49.3 $\pm$ 0.2      &  \textbf{51.3 $\pm$ 0.5}   \\ 
                                           \midrule                      
            \multirow{3}{*}{ImageNet-100}  & 10  & 9.5 $\pm$ 0.4                    & \textbf{36.0 $\pm$ 0.3}       & 28.2 $\pm$ 1.5         & 6.4 $\pm$ 0.1                   & \textbf{33.9 $\pm$ 0.1}      & 22.4 $\pm$ 3.3           & \textbf{23.6 $\pm$ 0.7}  &  {15.8 $\pm$ 0.2}   \\
                                           & 50  & 27.0 $\pm$ 0.4                   & 61.6 $\pm$ 0.1                & \textbf{67.9 $\pm$ 0.6}& 25.7 $\pm$ 0.3                  & 66.0 $\pm$ 0.6               & \textbf{70.8 $\pm$ 2.3}  & 51.5 $\pm$ 0.8  & \textbf{55.0 $\pm$ 1.8}   \\ 
                                           & 100 & -                                & 74.5 $\pm$ 0.4                & \textbf{75.1 $\pm$ 0.2}& -                               & 73.5 $\pm$ 0.8               & \textbf{77.6 $\pm$ 1.8}  &    70.8 $\pm$ 1.1   &  \textbf{76.7 $\pm$ 0.3}   \\ \midrule
           \multirow{3}{*}{ImageNet-1K}    & 10  & 21.3 $\pm$ 0.6                   & 42.0 $\pm$ 0.1                & \textbf{46.1 $\pm$ 0.4}& 30.9 $\pm$ 0.1                  & 48.3 $\pm$ 1.0               & \textbf{48.5 $\pm$ 1.6}  &    {33.1 $\pm$ 1.2}    &   {\textbf{35.5 $\pm$ 0.7}}   \\
                                           & 50  & 46.8 $\pm$ 0.2                   & 56.5 $\pm$ 0.1                & \textbf{59.2 $\pm$ 0.4}& 60.8 $\pm$ 0.5                  & 61.2 $\pm$ 0.4               & \textbf{66.1 $\pm$ 0.5}  &     52.8 $\pm$ 0.4      &   \textbf{56.2 $\pm$ 0.3}   \\ 
                                           & 100 & 52.8 $\pm$ 0.3                   & 59.8 $\pm$ 0.1                & \textbf{62.4 $\pm$ 0.2}& 62.8 $\pm$ 0.2                  & -                            & \textbf{67.6 $\pm$ 0.3}  &        56.2 $\pm$ 0.1   & \textbf{58.9 $\pm$ 0.3}   \\
                                           \bottomrule[1.4pt]
        \end{tabular}
    }
    \vspace{-0.8em}
    \caption{
        Comparison with SOTA dataset distillation methods using relatively large-scale backbones on five benchmarks across different scales. MobileNet-v2 is modified to match the low resolutions of CIFAR-10 and Tiny-ImageNet following~\cite{zhao2022decoupled}. Due to the limited table space, some prior methods that are slightly weaker or comparable with RDED are not listed, such as CDA and G-VBSM. Since IPC = 1 is not applicable to use \texttt{EarlyLate} strategy, thus under IPC = 1 setting, the single image in each class is optimized with a constant iteration.}
    \vspace{-.7em}
    \label{tb:resnet}
\end{table*}

\begin{table*}[t]
    \centering
    \scalebox{0.82}{
        \begin{tabular}{@{}cc|cccccc@{}c}
            \toprule
                                           &     & \multicolumn{6}{c}{ConvNet}                                                                                                        \\
                                           \cmidrule(lr){3-8}
            Dataset                        & IPC &      MTT~\cite{cazenavette2022dataset}            & IDM~\cite{zhao2023improved}        &    TESLA~\cite{cui2023scaling}            & DATM~\cite{guo2024lossless}                    & RDED~\cite{sun2024diversity}     &    DELT (Ours)                   \\  \midrule
            \multirow{3}{*}{Tiny-ImageNet} & 1   & 8.8 $\pm$ 0.3                                     & 10.1 $\pm$ 0.2                     & -                                         & \textbf{17.1 $\pm$ 0.3}                        & 12.0 $\pm$ 0.1                   & 12.4 $\pm$ 0.8          \\
                                           & 10  & 23.2 $\pm$ 0.2                                    & 21.9 $\pm$ 0.3                     & -                                         & 31.1 $\pm$ 0.3                                 & 39.6 $\pm$ 0.1                   & \textbf{40.0 $\pm$ 0.4}   \\
                                           & 50  & 28.0 $\pm$ 0.3                                    & 27.7 $\pm$ 0.3                     & -                                         & 39.7 $\pm$ 0.3                                 & 47.6 $\pm$ 0.2                   & \textbf{48.6 $\pm$ 0.2} \\ 
                                           \midrule
            \multirow{3}{*}{ImageNet-100}  & 10  & -                                                 & 17.1 $\pm$ 0.6                     & -                                         & -                                              & \textbf{29.6 $\pm$ 0.1}          & 24.7 $\pm$ 1.5          \\
                                           & 50  & -                                                 & 26.3 $\pm$ 0.4                     & -                                         & -                                              & 50.2 $\pm$ 0.2                   & \textbf{51.9 $\pm$ 1.1} \\ 
                                           & 100 & -                                                 & -                                  & -                                         & -                                              & 58.6 $\pm$ 0.4                   & \textbf{61.5 $\pm$ 0.5}\\ 
                                           \midrule
            \multirow{3}{*}{ImageNet-1K}   & 1 & -                                                   & -                                  &  7.7 $\pm$ 0.2                            & -                                              & 6.4 $\pm$ 0.1                    & \textbf{8.8 $\pm$ 0.5}  \\
                                           & 10  & -                                                 & -                                  &  17.8 $\pm$ 1.3                           & -                                              & 20.4 $\pm$ 0.1                   & \textbf{31.3 $\pm$ 0.8} \\
                                           & 50  & -                                                 & -                                  &  27.9 $\pm$ 1.2                           & -                                              & 38.4 $\pm$ 0.2                   & \textbf{41.7 $\pm$ 0.1} \\ 
                                           \bottomrule
        \end{tabular}
    }
    \vspace{-0.7em}
    \caption{
        {Comparison with SOTA dataset distillation methods using small-scale backbone architecture on three datasets.} {Following~\cite{cazenavette2022dataset,zhao2023improved,sun2024diversity}, Conv-4 for Tiny-ImageNet and ImageNet-1K, Conv-6 for ImageNet-100. Entries marked with ``-'' are missing due to scalability issue.}
    }
    \vspace{-1.5em}
    \label{tb:main}
\end{table*}

\subsection{Cross-architecture generalization}
An important characteristic of distilled datasets is their effectiveness in generalizing to novel training architectures. In this context, we assess the transferability of DELT's distilled datasets tailored for ImageNet-1K with {10} images per class. Following previous studies~\cite{yin2023squeeze,sun2024diversity}, we test our models using {five} distinct architectures: ResNet-18~\cite{he2016deep}, MobileNet-V2~\cite{sandler2018mobilenetv2}, MnasNet1\_3~\cite{tan2019mnasnet}, EfficientNet-B0~\cite{tan2019efficientnet}, and RegNet-Y-8GF~\cite{radosavovic2020designing}. As shown in Table~\ref{ablation_generalization}, our proposed approach demonstrates significantly better performance than other competitive methods on all these architectures. 

\begin{figure}[t]
  \centering
  \includegraphics[width=1.0\linewidth]{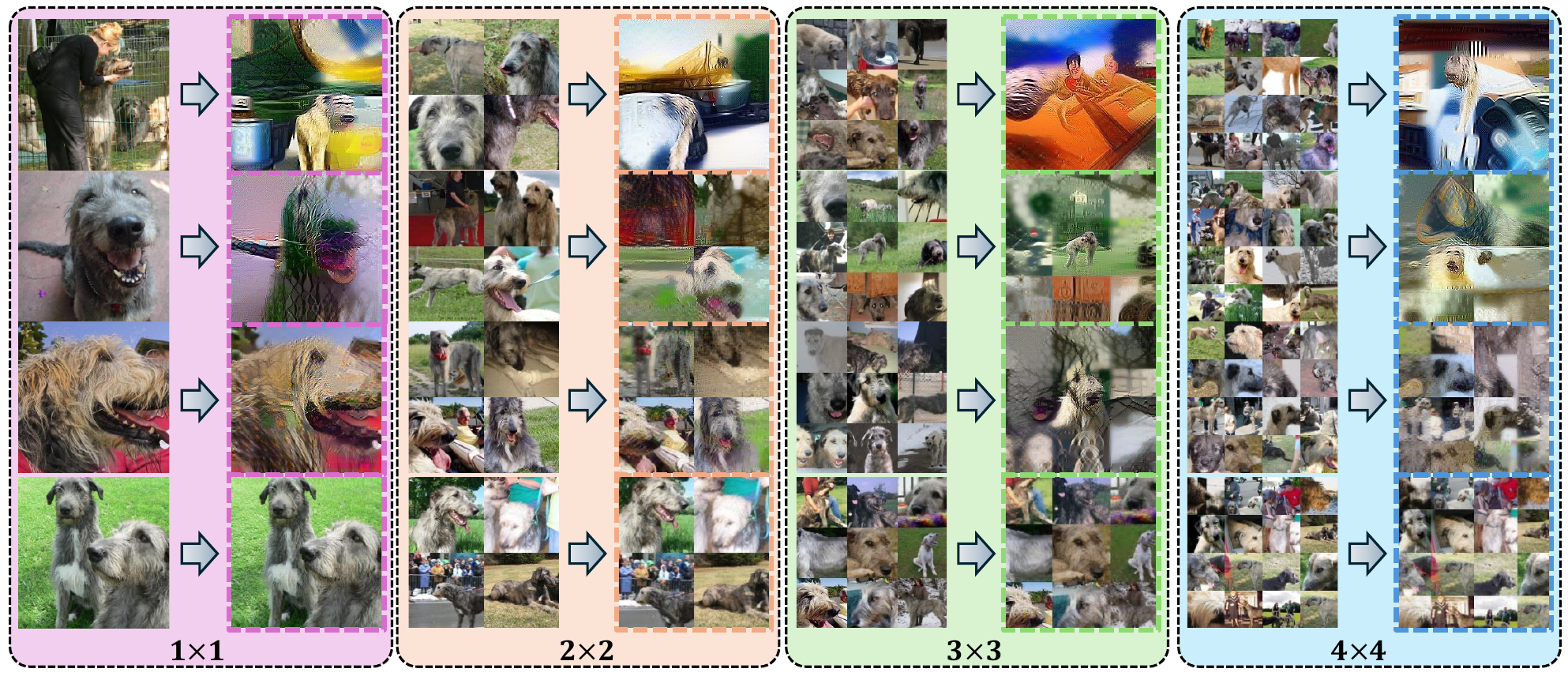}
  \vspace{-0.22in}
  \caption{Mosaic splicing patterns on ImageNet-1K using real image patches as the initialization. In each block, the left column is the starting real image initialized samples and right is the final optimized syntheses. From top to bottom are images generated by early training and late training.}
  \label{fig:vis_patterns}
   \vspace{-0.12in}
\end{figure}

\subsection{Ablation Study} \label{ablation}

\noindent{\bf Mosaic splicing pattern.} Mosaic stitching method~\cite{bochkovskiy2020yolov4} in RDED selects four crops from the train set as the optimal hyperparameter, and puts the contents of the four crops into a synthetic image that is directly used for post-validation. In this work, considering that we use different difficulty levels of selection for initialization, we examine different strategies of the Mosaic splicing patterns, including $1\times1$, $2\times2$, $3\times3$, $4\times4$, and $5\times5$ patches, as illustrated in Fig.~\ref{fig:vis_patterns}. The ablation results are shown in Table~\ref{ablation-num-patches}, it can be observed that $1\times1$ achieves the best accuracy.

\begin{table*}[h]
\centering
\resizebox{0.99\textwidth}{!}{
\begin{tabular}{@{}c|cccc}
\toprule
 Strategy &  SRe$^2$L~\cite{yin2023squeeze} w/o Init w/o \texttt{EarlyLate} &  SRe$^2$L~\cite{yin2023squeeze} w/o Init w/ \texttt{EarlyLate} & CDA~\cite{yin2023dataset} w/o Init w/o \texttt{EarlyLate} &  CDA~\cite{yin2023dataset} w/o Init  w/ \texttt{EarlyLate} \\
Acc. & 46.8  &  53.4$_{\textcolor{Aquamarine}{\bf (+6.6)}}$  &  53.5   &   \bf 55.9$_{\textcolor{Aquamarine}{\bf (+2.4)}}$ \\
 \bottomrule
\end{tabular}
}
\vspace{-0.1in}
\caption{Performance comparison without real image initialization on ImageNet-1K with IPC 50.}
\label{without_init}
\vspace{-0.05in}
\end{table*}

\begin{table*}[h]
    \centering
    \begin{minipage}{0.48 \linewidth}
        \centering
        \resizebox{0.38\textwidth}{!}{
        \begin{tabular}{lc}
            \toprule
            \# Patches     & Top 1 acc\\
            \midrule
            $1 \times 1$ &  \textbf{57.57} \\
            $2 \times 2$ & 56.92  \\
            $3 \times 3$ & 56.62  \\
            $4 \times 4$ & 56.71  \\
            $5 \times 5$ & 56.51  \\
            \bottomrule
        \end{tabular}
        }
        \subcaption{\textbf{Number of patches}. Ablation on initializing different numbers of scoring patches. Results are from ResNet-18 on ImageNet-1K for 500 iterations to synthesize 50 IPCs. Our optimization-based method favors $1 \times 1$ initialized patch, and will involve inconsistency and noise using more objects.}
        \label{ablation-num-patches}
    \end{minipage}
    \hfill
    \begin{minipage}{0.48 \linewidth}
        \centering
        \vspace{-0.55in}
        \resizebox{0.52\textwidth}{!}{
        \begin{tabular}{lc}
            \toprule
            Selection criteria     & Top 1 acc    \\
            \midrule
            Lowest probability & 57.55 \\
            Medium probability & \textbf{57.67}  \\
            Highest probability & 57.03  \\
            \bottomrule
        \end{tabular}
        }
        \subcaption{\textbf{Selection criteria}. Initializing $1\times 1$ images selected according to teacher model's probability}
        \label{ablation-selection-criteria}
    \end{minipage}
    \begin{minipage}{0.5 \linewidth}
        \centering
        \resizebox{0.47\textwidth}{!}{
        \begin{tabular}{ccc}
            \toprule
            Iterations &\multicolumn{2}{c}{Round Iterations (RI)}\\
              (MI)   & 500 & 1K  \\
            \midrule
            1K & 44.87 & 43.71 \\
            2K & 45.61 & 44.40\\
            4K & \textbf{46.42} & 44.66\\
            \bottomrule
        \end{tabular}
        }
        \subcaption{\textbf{Round Iterations}. Top-1 acc. of our method for IPC 10 using different round iterations with ResNet-18.}
        \label{ablation-round-iterations}
    \end{minipage}
    \begin{minipage}{0.48 \linewidth}
        \vspace{-0.45in}
        \centering
        \resizebox{0.72\textwidth}{!}{
        \begin{tabular}{ccc}
            \toprule
             Dataset  & CDA~\cite{yin2023dataset} + Our init. & Ours  \\
            \midrule
            ImageNet-1K  & 43.5 & \bf46.1 \\   
            Tiny-ImageNet  & 42.2  & \bf43.0  \\  
            CIFAR-10    & 39.4 &  \bf 43.0 \\
            \bottomrule
        \end{tabular}
        }
        \subcaption{Ablation on init. and \texttt{EarlyLate} under IPC 10.}
        \label{ablation_CDA}
                \centering
        \resizebox{0.85\textwidth}{!}{
        \begin{tabular}{cccc}
            \toprule
             IPC & RDED~\cite{sun2024diversity} & MinimaxDiffusion~\cite{gu2024efficient} & Ours  \\
            \midrule
            10 & 42.0  & 44.3   & \bf 46.1 \\
            50 & 56.5  &  58.6 & \bf 59.2  \\
            \bottomrule
        \end{tabular}
        }
        \subcaption{Comparison with real and diffusion-generated data.}
        \label{ablation_diffusion}
    \end{minipage}
    \vspace{-0.09in}
    \caption{\textbf{Ablation experiments} on various aspects of our framework with ResNet-18 on ImageNet-1K.}
    \label{init-ablations}
        \centering
    \vspace{0.06in}
     \resizebox{0.87\textwidth}{!}{
\begin{tabular}{@{}clccccc@{}}
\toprule
\multicolumn{2}{c}{Recover / Validation}                            & ResNet-18 & EfficientNet-B0 & MobileNet-V2 & MnasNet1\_3 & RegNet-Y-8GF \\ \midrule
\multirow{5}{*}{ResNet-18} & \multicolumn{1}{l|}{SRe$^2$L~\cite{yin2023squeeze}$^\dagger$} & {41.9}      & 41.9            & 33.1         & 39.3        & 51.5         \\
 & \multicolumn{1}{l|}{CDA~\cite{yin2023dataset}}         & {42.2} & 43.9 & 34.2 & 39.7 & 52.9 \\
 & \multicolumn{1}{l|}{G-VBSM~\cite{shao2023generalized}} & {41.4} & 42.6 & 33.5 & 40.1 & 52.2 \\
 & \multicolumn{1}{l|}{RDED~\cite{sun2024diversity}}      & 42.3 & 42.8 & 34.4 & 40.0 & 54.8 \\
 & \multicolumn{1}{l|}{\bf Ours}                       &  \ \ \qquad\textbf{46.4$_{\textcolor{Aquamarine}{\bf (+4.1)}}$}    & \ \ \qquad\textbf{47.1$_{\textcolor{Aquamarine}{\bf (+4.3)}}$}     & \ \ \qquad\textbf{36.1$_{\textcolor{Aquamarine}{\bf (+1.7)}}$}      & \ \ \qquad\textbf{40.7$_{\textcolor{Aquamarine}{\bf (+0.7)}}$}    &  \ \ \qquad\textbf{57.5$_{\textcolor{Aquamarine}{\bf (+2.7)}}$}    \\ \bottomrule
\end{tabular}
}
\vspace{-0.05in}
\caption{\textbf{Cross-architecture generalization}. Results are evaluated on IPC 10. $^\dagger$ is reproduced following CDA's configuration.}
\label{ablation_generalization}
\vspace{-0.2in}
\end{table*}

\noindent{\bf Initialization.} We examine how different initialization strategies affect final performance, including: choosing the lowest probability crops, medium probability crops and highest probability crops. Our results are shown in Table~\ref{ablation-selection-criteria}. Overall, the performance gap between different strategies is not significant, and selecting the medium probability crops as the initialization achieves the best accuracy.
\\
\noindent{\bf Optimization iterations.} We examine two types of optimization iterations: maximum iteration (MI) for the earliest batch training and round iteration (RI) as the iteration gap between the two adjacent batches. 
As shown in Table~\ref{ablation-round-iterations}, we test MI values of 1K, 2K, and 4K, using 500 and 1K iterations for each RI. Note that when MI is set to 1K, it is not feasible to use 1K as RI. The results show that 4K (same as~\cite{yin2023squeeze,yin2023dataset}) MI and 500 RI achieves the best accuracy.
\\
\noindent{\bf Early-only vs. EarlyLate.}  Early-only is equivalent to using constant MI to optimize each image. This will transform to baseline {\em batch-to-global} matching of CDA~\cite{yin2023dataset} + real image initialization. Our results in Table~\ref{ablation_CDA} clearly show that the \texttt{EarlyLate} training brings a significant improvement on final performance. More importantly, this strategy is the key factor in enhancing generation diversity.

\noindent{\bf Real image stitching vs. Minimax diffusion vs. Ours.} We further compare our approach with real image stitching~\cite{sun2024diversity} and diffusion generation~\cite{gu2024efficient}. The results are presented in Table~\ref{ablation_diffusion}. While the first two methods produce more realistic images, each image contains limited information. In contrast, our method achieves the best final performance.

\noindent {\bf Without real image initialization.} Our \texttt{EarlyLate} strategy enhances the performance of 1$\sim$2\% over the initialization as shown in Table~\ref{ablation_CDA}. Without initialization, our method consistently improves by 2.4\% as shown in Table~\ref{without_init}.

\vspace{-0.05in}
\subsection{Computational Analysis}

For image optimization-based methods like SRe$^2$L and CDA, the total computational cost is calculated as $N\times T$, where $N$ is the MI. In our \texttt{EarlyLate} scheme, the first batch of images undergo $T_1$ iterations (where $T_1=T$). Subsequent batches are processed with progressively fewer iterations, such as $T_2$ ($T_2=T_1-\text{RI}$) for the next set, and so forth. The iterations for the final batch are reduced to {RI} which is $1/j$ of the standard count (where $j=$ 4 or 8 in our ablation),  the total number of our optimization iterations required is $N\times T-\frac{j(j-1)}{2}\text{RI}$, which is roughly $2/3$ of prior {\em batch-to-global} matching methods. Our real time consumptions for data generation are shown in Table~\ref{table:computation}, note that the smaller the dataset like CIFAR, the more time is spent on loading and processing the data, rather than training.

\subsection{Visualization of DELT}

\begin{figure*}[t]
  \centering
  \vspace{-0.17in}
  \includegraphics[width=0.82\linewidth]{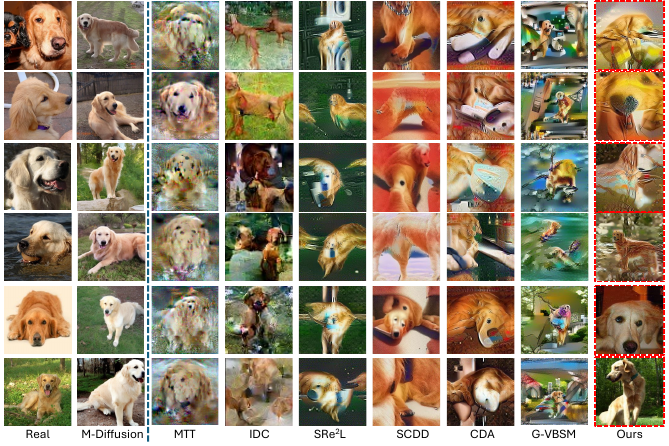}
  \vspace{-0.13in}
  \caption{Distilled dataset visualization compared with other image optimization-based methods.}
  \label{fig:vis_all}
   \vspace{-0.05in}
\end{figure*}

Fig.~\ref{fig:vis_all} illustrates a comprehensive visual comparison between randomly selected synthetic images from our distilled dataset and those from the real image patches~\cite{sun2024diversity}, MinimaxDiffusion~\cite{gu2024efficient}, MTT~\cite{cazenavette2022dataset}, IDC~\cite{pmlr-v162-kim22c}, SRe$^2$L~\cite{yin2023squeeze}, SCDD~\cite{zhou2024self}, CDA~\cite{yin2023dataset} and G-VBSM~\cite{shao2023generalized} distilled data. It can be observed that the images generated by each method have their own characteristics. MinimaxDiffusion leverages the diffusion model to synthesize images which is close to the real ones. However, as in our above ablation, both real and diffusion-generated data are inferior to ours. MTT results show noticeable artifacts and distortions, the objects in all images are located in the middle of the generations, the diversity is limited. IDC results also show distorted and less recognizable dog images, but diversity is increased. SRe$^2$L exhibits some dog features but with significant distortions and a similar simple background. SCDD shows more recognizable dog features but still the color is simple and monochromatic, the same situation happens in CDA. G-VBSM shows more colorful patterns, possibly due to recovery from multiple different networks, but all generations are in the same pattern and the diversity is not large. Our approach's synthetic images exhibit a higher degree of diversity, including both compressed distorted images from long-optimized initializations and clear, recognizable dog images from short-optimized initializations, a unique capability not present in other methods.

\begin{table*}[h]
    \centering
    \resizebox{0.63\textwidth}{!}{
    \begin{tabular}{lccc}
      \toprule
      & \multicolumn{3}{c}{Dataset (hours) under same 4K iterations for all methods}                   \\
      \cmidrule(r){2-4}
      Method                             & ImageNet-1K     & Tiny-ImageNet &\textcolor{gray}{CIFAR-10} \\
      \midrule
      G-VBSM~\cite{shao2023generalized}  & 114.1             & 5.5          &  \textcolor{gray}{0.195} \\
      SRe$^2$L~\cite{yin2023squeeze}     &  29.0           &  5.0        &  \textcolor{gray}{0.084} \\
      CDA~\cite{yin2023dataset}          &  29.0            &  5.0        & \textcolor{gray}{0.084}  \\
      Ours ({RI} = 1K)               & \qquad\quad \bf 18.8$_{\textcolor{Aquamarine}{\bf (\downarrow35.2\%)}}$            & \qquad\quad \bf 3.6$_{\textcolor{Aquamarine}{\bf (\downarrow28.0\%)}}$        &   \qquad\quad \textcolor{gray}{\bf 0.084$_{\textcolor{Aquamarine}{\bf (\downarrow0.0\%)}}$} \\
      Ours ({RI} = 500)              & \qquad\quad \bf 17.6$_{\textcolor{Aquamarine}{\bf (\downarrow39.3\%)}}$        & \qquad\quad \bf 3.4$_{\textcolor{Aquamarine}{\bf (\downarrow32.0\%)}}$        &   \qquad\quad \textcolor{gray}{\bf 0.083$_{\textcolor{Aquamarine}{\bf (\downarrow1.1\%)}}$}  \\
      \bottomrule
    \end{tabular}
    }
    \vspace{-0.06in}
    \caption{{\bf Actual computational consumption} (hours under IPC 50) in data synthesis with image optimization-based methods on a single NVIDIA 4090 GPU. A total 4K iterations are used for all methods and datasets to ensure fair comparisons. ``{{RI}}'' represents {\em round iterations}.}
    \label{table:computation}
    \vspace{-0.14in}
  \end{table*}

\subsection{Application I: Data-free Network Pruning}

Our distilled dataset acts as a multifunctional training tool and boosts the adaptability for diverse downstream applications. We validate its utility in the scenario of data-free network pruning~\cite{srinivas2015data}. Table~\ref{pruning} shows the applicability of our dataset in this task when pruning 50\% weights, where it significantly surpasses previous methods such as SRe$^2$L and RDED under IPC 10 and 50.

\begin{table}[h]
\centering
\resizebox{0.35\textwidth}{!}{
\begin{tabular}{c|ccl}
\toprule
       & SRe$^2$L~\cite{yin2023squeeze} & RDED~\cite{sun2024diversity} & Ours \\ \midrule
IPC 10 & 12.5  & 13.2 & \bf 17.9$_{\textcolor{Aquamarine}{\bf (+4.7)}}$ \\ 
IPC 50 & 31.7  & 42.8 & \bf 44.8$_{\textcolor{Aquamarine}{\bf (+2.0)}}$ \\ \bottomrule
\end{tabular}
}
\vspace{-0.1in}
\caption{Accuracy of data-free network pruning using slimming~\cite{liu2017learning} on VGG11-BN~\cite{simonyan2014very}.}
\label{pruning}
\vspace{-0.15in}
\end{table}

\subsection{Application II: Continual Learning}

\begin{figure}
  \centering
    \includegraphics[width=0.8\linewidth]{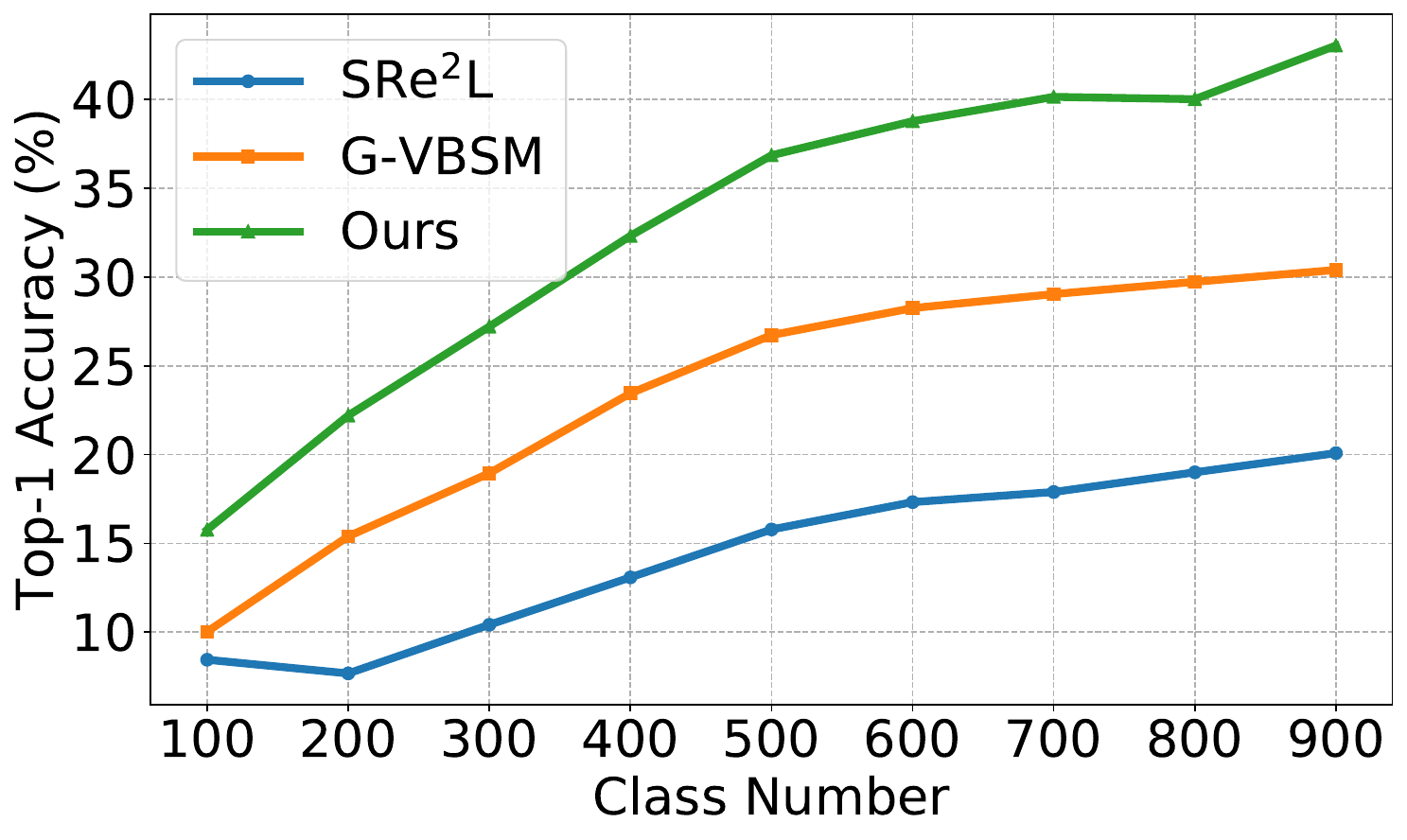}
  \vspace{-0.15in}
  \caption{Continual learning results.}
  \label{fig:CL}
   \vspace{-0.25in}
\end{figure}

 We examine the effectiveness of DELT generated images in the continual learning scenario. Following the setup in prior studies~\cite{yin2023squeeze,DBLP:conf/wacv/ZhaoB23}, we perform 100-step class-incremental experiments on ImageNet-1K, comparing our results with the baselines G-VBSM and SRe$^2$L. As shown in Fig.~\ref{fig:CL}, our DELT distilled dataset significantly outperforms G-VBSM, with an average improvement of about 10\% on the 100-step class-incremental learning task. This highlights the significant benefits of deploying DELT, particularly in mitigating the challenges of continual learning.

\vspace{-0.05in}
\section{Conclusion}
\label{sec:conclusion}

We have introduced a new training strategy, {\texttt{EarlyLate}}, to improve image diversity in {\em batch-to-global} matching scenarios for dataset distillation. The proposed approach organizes predefined IPC samples into smaller, manageable subtasks and utilizes local optimizations. This strategy helps in refining each subset into distributions characteristic of different phases, thereby mitigating the homogeneity typically caused by a singular optimization process. The images refined through this method exhibit robust generalization across the entire task. We have extensively evaluated this approach on CIFAR-10, Tiny-ImageNet, ImageNet-1K, and its variants. Our empirical findings indicate that our approach significantly outperforms prior state-of-the-art methods across various IPC configurations.

\section*{Acknowledgments}
This research is supported by the MBZUAI-WIS Joint Program for AI Research and the Google Research award grant.
{
    \small
    \bibliographystyle{ieeenat_fullname}
    \bibliography{main}
}
\clearpage
\appendix
\section*{\Large Appendix}
\label{sec:appendix_section}

\section{Limitations} \label{BI}
Our method effectively avoids the issue of insufficient data diversity generated by {\em batch-to-global} methods and reduces the computational cost of the generation process. However, there is still a performance gap when training the model on our generated data compared to training on the original dataset. Also, our short-optimized data exhibits similar appearance and semantic information to the original images, which has demonstrated better privacy protection than prior train-free methods but may still potentially leak the privacy of the original dataset to some extent.

\section{More Training Details}

\begin{table*}[h]
        \centering
        \begin{subtable}[t]{0.49\textwidth}
            \centering
            \caption{Validation settings}
            \resizebox{0.76\textwidth}{!}{
            \begin{tabular}{@{}l|l@{}}
            config                              & value            \\ \midrule
            optimizer                           & AdamW            \\
            \multirow{2}{*}{base learning rate} & 0.001 (all)      \\
                                                & 0.0025 (MobileNet-v2)\\
            weight decay                        & 0.01             \\
            \multirow{3}{*}{batch size}         & 100 (IPC 50)      \\
                                                & 50 (IPC 10)       \\
                                                & 10 (IPC 1)       \\
            learning rate schedule              & cosine decay     \\
            training epoch                      & 300              \\
            \multirow{3}{*}{augmentation}       & RandAugment   \\
                                                & RandomResizedCrop   \\
                                                & RandomHorizontalFlip   \\
            \end{tabular}
            }
            \label{tab:config_validation}
        \end{subtable}
        \begin{subtable}[t]{0.49\textwidth}
            \centering
            \caption{Recovery settings}
            \begin{tabular}{@{}l|l@{}}
            config                 & value                           \\ \midrule
            $\alpha_{\text{BN}}$   & 0.01                            \\
            optimizer              & Adam                            \\
            base learning rate     & 0.25                          \\
            momentum     & $\beta_1$, $\beta_2$ = 0.5, 0.9 \\
            batch size             & 100                             \\
            learning rate schedule & cosine decay                \\
            recovery iteration     & 4,000     \\
            round iteration        & 500     [IPC 10, 50, 100]\\
            initialization         & top medium     \\
            augmentation           & RandomResizedCrop              
            \end{tabular}
            \label{tab:config_recover}
        \end{subtable}
        \begin{subtable}[t]{0.99\textwidth}
            \centering
            \caption{Dataset-specific settings in recovery}
            \scalebox{0.83}{
                \begin{tabular}{@{}l|ccccc@{}}
                config                                   & CIFAR10       & Tiny-ImageNet  & ImageNette    & ImageNet-100  & ImageNet-1K   \\ \midrule
                RandAugment ($\texttt{m}$)               & 5             & 4              & 6             & 6             & 6             \\
                RandAugment ($\texttt{n}$)               & 4             & 3              & 2             & 2             & 2             \\
                RandAugment ($\texttt{mstd}$)            & 1.0           & 1.0            & 1.0           & 1.0           & 1.0           \\
                \multirow{4}{*}{IPC1 Recovery Iterations}& 2K (R18) & 500 (R18) & 1K (R18) & -             & 3K (Conv4)    \\ 
                                                         & 3K (R101)& 500 (R101)& 1K (R101)& -             & -             \\ 
                                                         & 2K (MobileNet)& 500 (MobileNet)& 2K (MobileNet)& -             & -             \\ 
                                                         & -             & 1K (Conv4)     & 4K (Conv5)    & -             & -             \\ 
                \end{tabular}
            }
            \label{tab:config_dataset}
        \end{subtable}
    \caption{Hyper-parameter settings.}
    \label{tab:config}
    \vspace{1em}

\centering
\resizebox{0.95\textwidth}{!}{
\begin{tabular}{@{}c|ccc}
\toprule
 Initialization &  SRe$^2$L + w/ Init w/o \texttt{EarlyLate} & CDA + Init w/o \texttt{EarlyLate} &  CDA + w/ Init + w/ \texttt{EarlyLate} (Ours) \\
    2$\times$2      &  55.3    & 56.9     &  \bf 58.2$_{\textcolor{Aquamarine}{\bf (+1.3)}}$  \\
    3$\times$3      &  55.8    & 56.6     &  \bf 58.1$_{\textcolor{Aquamarine}{\bf (+1.5)}}$ \\
    4$\times$4      &  55.2    & 56.7     &  \bf 57.4$_{\textcolor{Aquamarine}{\bf (+0.7)}}$ \\
    5$\times$5      &  54.6    & 56.5     &  \bf 57.3$_{\textcolor{Aquamarine}{\bf (+0.8)}}$ \\
\bottomrule
\end{tabular}
}
\caption{Performance comparison w/ and w/o \texttt{EarlyLate} on ImageNet-1K under IPC 50.}
\label{ablation_init}
\end{table*}

For reproducibility, we provide all our hyperparameter settings used in our experiments in Table \ref{tab:config}, we outline such details below.

\noindent{\bf Squeezing and Pre-trained models.} Following the previous works \cite{yin2023squeeze,yin2023dataset}, we use the official PyTorch \cite{paszke2019pytorch} pre-trained ResNet-18 model for ImageNet-1K, and we use the same official Torchvision \cite{paszke2019pytorch} code to obtain our pre-trained models, ResNet-18 and ConvNet, for the other datasets.

\noindent{\bf Ranking.} For our initialization, we simply use ResNet-18 pre-trained models to rank and select the medium images as initialization for all our datasets, except for ImageNet-100 where we simply extract the medium images based on the rankings of the original ImageNet-1K.

\begin{figure}[t]
  \centering
  \includegraphics[width=0.98\linewidth]{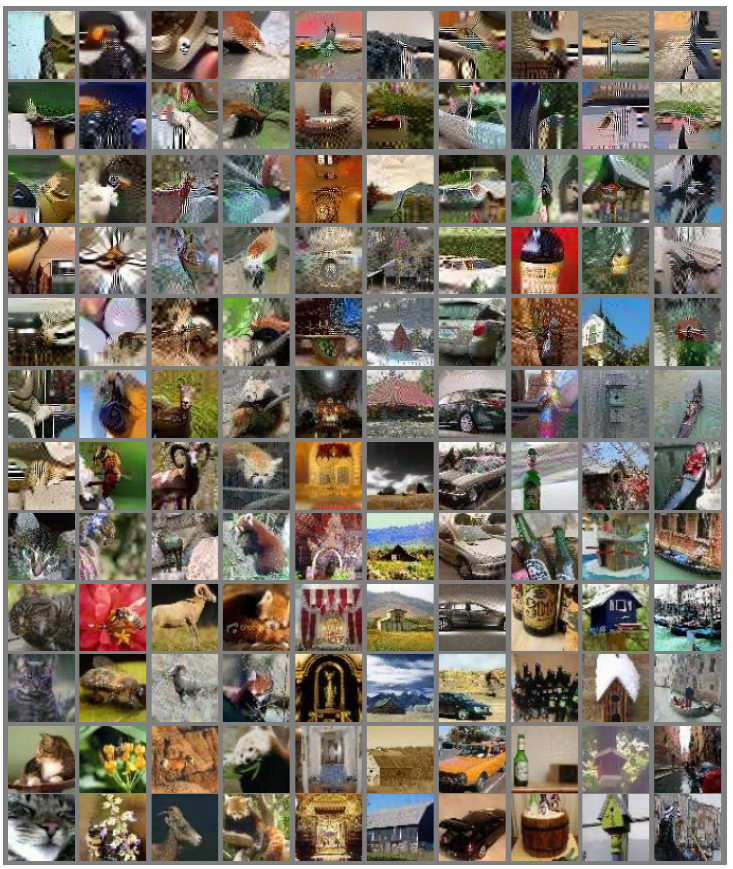}
  \vspace{-0.1in}
  \caption{Synthetic image visualizations on Tiny-ImageNet generated by our DELT. }
  \label{fig:vis_patterns1}
   \vspace{-0.1in}
\end{figure}

\begin{figure}[t]
  \centering
  \includegraphics[width=0.98\linewidth]{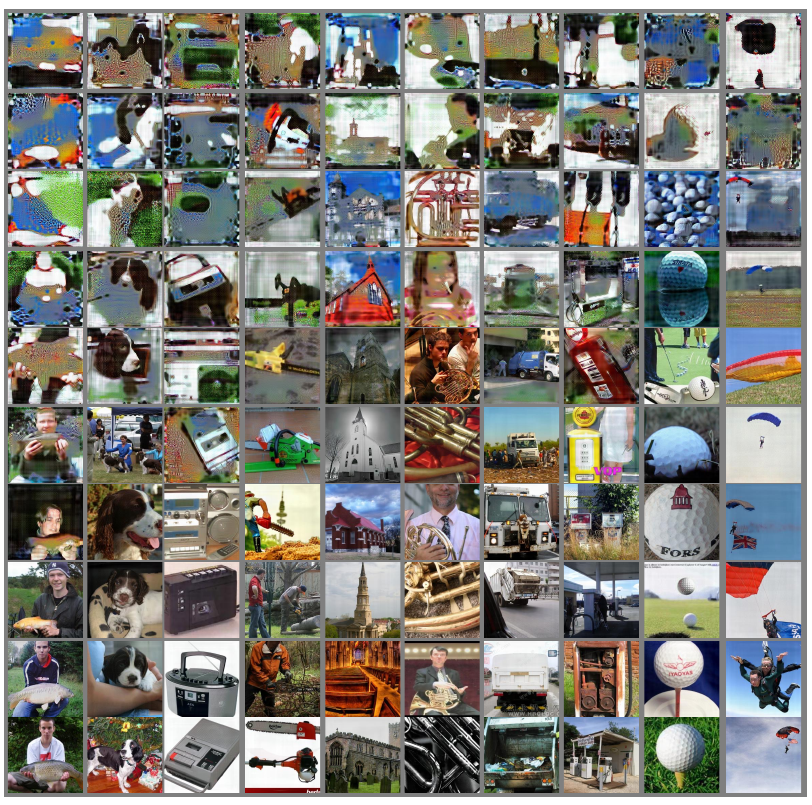}
  \vspace{-0.1in}
  \caption{Synthetic image visualizations on ImageNette generated by our DELT. }
  \label{fig:vis_patterns2}
   \vspace{-0.1in}
\end{figure}

\begin{figure}[t]
  \centering
  \includegraphics[width=0.98\linewidth]{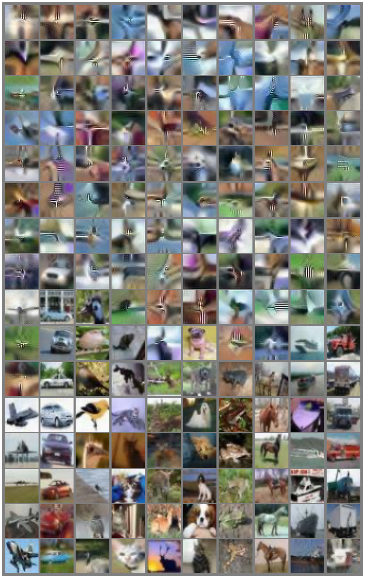}
  \vspace{-0.1in}
  \caption{Synthetic images on CIFAR-10 generated by our DELT.}
  \label{fig:vis_patterns3}
   \vspace{-0.1in}
\end{figure}

\begin{table}[h]
\centering
\resizebox{0.17\textwidth}{!}{
\begin{tabular}{@{}c|c}
\toprule
Order	& DELT \\
Random	& 67.9 \\
Ascending &	67.2 \\
Descending	& 67.7 \\
Our DELT  &	68.2 \\
\bottomrule
\end{tabular}
}
\caption{Impact of using different ordering on ImageNet-100 when having the same initialized images of the median probability.}
\label{order1}
\end{table}

\begin{table}[h]
\centering
\resizebox{0.2\textwidth}{!}{
\begin{tabular}{@{}c|c}
\toprule
Selection Strategy &	DELT \\
Random	&  67.7  \\
Ascending	&  66.9 \\
Descending	&  67.3 \\
Our DELT	&  68.2 \\
\bottomrule
\end{tabular}
}
\caption{Comparison of the performance of different initialization strategies. The initialized images are different.}
\label{order2}
\end{table}

\noindent{\bf Recovery.} For our synthetic stage, we provide the details of general hyperparameters used for different datasets, including ImageNet-1K, ImageNet-100, ImageNette, Tiny-ImageNet, and CIFAR10, in Table \ref{tab:config_recover}. Synthesizing a single image per class, i.e., IPC = 1, is special as we cannot use {\em rounds}, so we apply individual numbers of iterations based on both the dataset scale and the validation teacher model as outlined in Table \ref{tab:config_dataset}.
We also utilize the BatchNorm distribution regularization term as in SRe$^2$L~\cite{yin2023squeeze} for Eq. 7 in the main paper to improve the quality of the generated images:
\begin{equation} \label{batch_to_global}
\footnotesize
\mathcal{R}_{\text {reg }}(\tilde{\boldsymbol{x}})=\sum_l\left|\mu_l(\widetilde{\boldsymbol{x}})-\mathbf{B N}_l^{\mathrm{RM}}\right|_2+\sum_l\left|\sigma_l^2(\tilde{\boldsymbol{x}})-\mathbf{B N}_l^{\mathrm{RV}}\right|_2
\end{equation}
where $l$ is the index of BN layer, $\mu_l(\widetilde{\boldsymbol{x}})$ and $\sigma_l^2(\widetilde{\boldsymbol{x}})$ are mean and variance. $\mathbf{B N}_l^{\mathrm{RM}}$ and $\mathbf{B N}_l^{\mathrm{RV}}$ are running mean and running variance in pre-trained model at $l$-th layer, which are globally counted.

\noindent{\bf Validation.} This includes the soft-label generation~\cite{shen2022fast} as used in SRe$^2$L, post-training and evaluation. We outline such details in Table \ref{tab:config_validation}. We use $\texttt{timm}$'s version of RandAugment \cite{cubuk2020randaugment} with different settings depending on the synthesized dataset being validated, as shown in Table \ref{tab:config_dataset}.

\section{More Visualization}

We provide more visualizations on synthetic Tiny-ImageNet, ImageNette and CIFAR-10 datasets in Fig.~\ref{fig:vis_patterns1}, \ref{fig:vis_patterns2}, \ref{fig:vis_patterns3}. In each figure, each column represents a different class, with images progressing from long optimization at the top to short optimization at the bottom.

\section{More Ablation}

\noindent {\bf Performance comparison w/ and w/o \texttt{EarlyLate}.} We compare using the recent CDA and SRe$^2$L as the base frameworks. The performance comparison for IPC 50 on ImageNet-1K is shown in Table~\ref{ablation_init}. It can be observed that the proposed {\em EarlyLate} strategy enhances the performance by around 1\% with the initialization.

\noindent {\bf Comparison of random, ascending and descending orders by patch probability.} In our DELT, we select the $N$ patches with scores around the median from the teacher, where the score represents the probability of the true class. To order them, we start with the median, and we go back and forth expanding the window around the median until we cover the number of IPCs, refer to Fig.~\ref{fig:selection} of the main paper for details. The rationale is that these patches present a medium difficulty level for the teacher, allowing more potential for information enhancement through distillation gradients while having a good starting point of information. We empirically validate it by comparing different strategies in Table 4b of the main paper.

As shown in Table~\ref{order1}, we present the impact of using different orderings on ImageNet-100 when having the same initialized images, those around the median.
We also include a comparison of different initialization strategies based on the order in Table~\ref{order2}. Unlike Table~\ref{order1}, the initialized images here are different across different strategies.

\noindent{\bf Different initial crop ranges in random crop augmentation for our DELT method.} Table~\ref{crop} compares different initial crop ranges in random crop augmentation for our DELT method. As shown in the results, the 0.08-1.0 range yields the best performance, which is the ablation and support for the default setting in our framework.

\begin{table}[h]
\centering
\resizebox{0.3\textwidth}{!}{
\begin{tabular}{@{}cc}
Random Crop Range &	Top 1-acc \\
\toprule
0.08-1.0 &	\bf 67.8  \\
0.2-1.0	& 67.3  \\
0.5-1.0	& 66.3  \\
0.8-1.0	& 66.3  \\
 \bottomrule
\end{tabular}
}
\vspace{-0.05in}
\caption{Different initial crop ranges in random crop augmentation for our DELT method.}
\label{crop}
\vspace{-0.1in}
\end{table}

\begin{figure}[t]
  \centering
  \includegraphics[width=.9\linewidth]{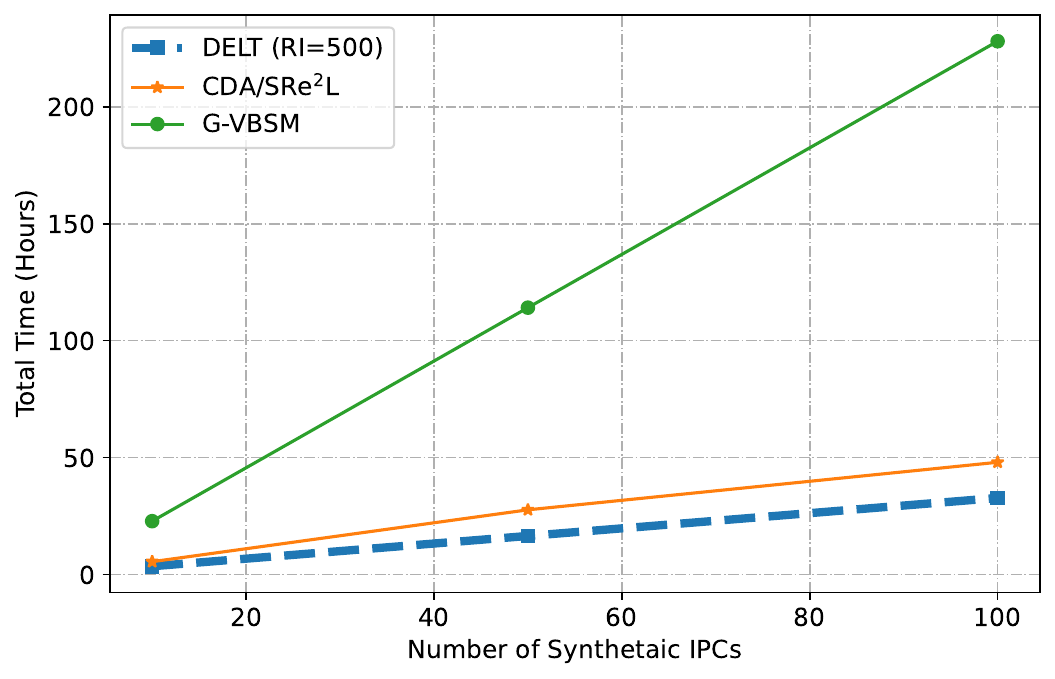}
   \includegraphics[width=.9\linewidth]{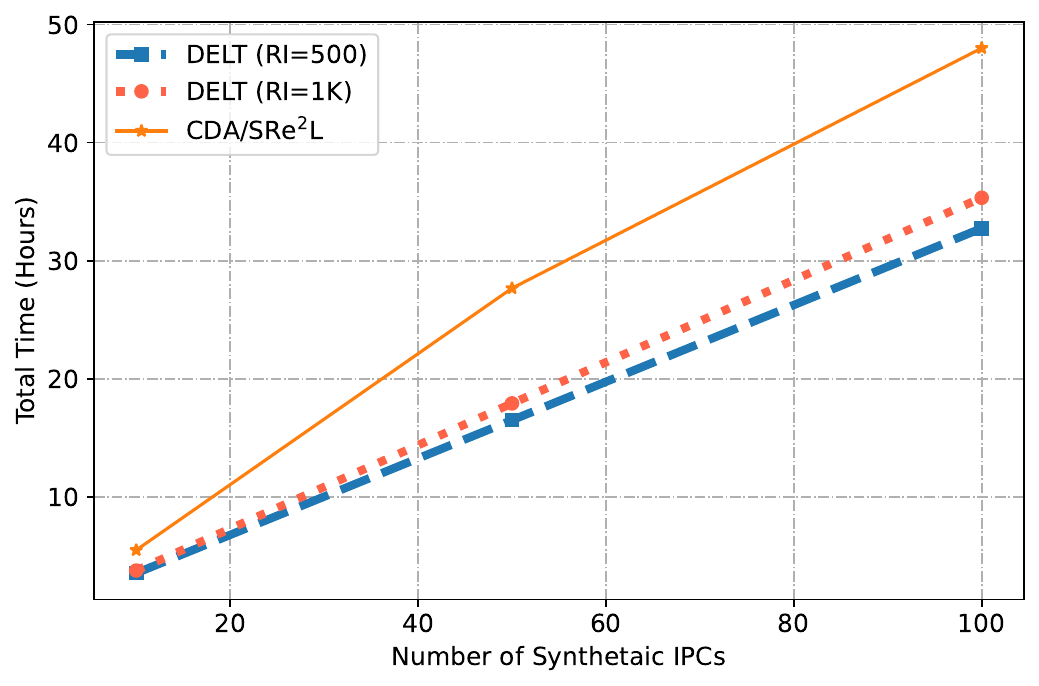}
  \vspace{-0.1in}
  \caption{Visualization of computation time consumption on our DELT and other methods including CDA, SRe$^2$L, and G-VBSM.}
  \label{fig:comp}
   \vspace{-0.1in}
\end{figure}

\section{Computational Efficiency}

Fig.~\ref{fig:comp} illustrates the computation time required for our DELT method compared to other methods, including CDA, SRe$^2$L, and G-VBSM, at various numbers of IPCs. The top subfigure shows a comparison of DELT (with a RI of 500), CDA/SRe$^2$L, and G-VBSM, where G-VBSM demonstrates the highest computation time, scaling significantly as the number of IPCs increases. In contrast, both DELT and CDA/SRe$^2$L maintain relatively low and consistent computation times, with DELT slightly outperforming CDA/SRe$^2$L. The bottom subfigure further compares DELT with RIs of 500 and 1,000 against CDA/SRe$^2$L, highlighting that both configurations of DELT offer lower or comparable computation times to CDA/SRe$^2$L across IPC values, with minimal increase as the IPC count rises. These results emphasize DELT's efficiency in computation time, particularly in comparison to G-VBSM, making it a computationally efficient choice for scenarios with larger datasets.

\end{document}